\documentclass{article}

    \PassOptionsToPackage{numbers, compress}{natbib}
     \usepackage[final]{neurips_2020}

\usepackage[utf8]{inputenc} % allow utf-8 input
\usepackage[T1]{fontenc}    % use 8-bit T1 fonts
\usepackage{hyperref}       % hyperlinks
\usepackage{url}            % simple URL typesetting
\usepackage{booktabs}       % professional-quality tables
\usepackage{amsfonts}       % blackboard math symbols
\usepackage{nicefrac}       % compact symbols for 1/2, etc.
\usepackage{microtype}      % microtypography
\usepackage{graphicx}
\usepackage{caption}
\usepackage{subcaption}
\usepackage{natbib}
\usepackage{xcolor}
\title{Who is more ready to get back in shape?} 

\author{%
  Rajius Idzalika \\
  Pulse Lab Jakarta\\
  \texttt{rajius.idzalika@un.or.id} \\
}

\begin{document}

\maketitle

\begin{abstract}
This empirical study estimates resilience (adaptive capacity) around the periods of the 2013 heavy flood in Cambodia. We use nearly 1.2 million microfinance institution (MFI) customer data and implement the unsupervised learning method. Our results highlight the opportunity to develop resilience by having a better understanding of which areas are likely to be more or less resilient based on the characteristics of the MFI customers, and the individual choices or situations that support stronger adaptiveness. We also discuss the limitation of this approach. 
\end{abstract}

\section{Introduction}
\label{intro}
The progress of sustainable growth in many developing and low income countries are constantly challenged by stresses and shocks, including from the extreme climate changes and the recent COVID-19 pandemic. The capacity to adapt to those hardships is crucial to avoid the disruption of development achievements and future development trajectories. Resilience, as the positive results of such a successful adaptation \cite{manyena2011disaster}\cite{houston2015bouncing}, can contribute to overcome the negative consequences of tragic events. The existing literature around this topic is pretty limited, though, where the typical data collection is through survey. In a low supporting environment of quality data collection, finding and then reusing alternative data sources whose main purpose of collection is not about resilience, but some of their features are fairly relevant to resilience, could be an interesting option. This paper introduces the possibility of using financial data and deploying unsupervised learning to learn the dynamic of adaptive capacity in Cambodia, one of the most vulnerable low income countries in the world. This is a piece of work on resilience. A more comprehensive study would include other aspects such as access to basic services, social safety nets and coping strategies. We hope, nevertheless, our work inspires more resilience studies in countries with little resources by using the existing non-conventional data sets coupled with machine learning. 

\section{Context and Related Works}
\label{works}
Agriculture is a dominant sector in Cambodia with 35 percent of its contribution to total GDP and employs the majority of the population \cite{fao20}. Due to the geographical location, this country is highly prone to natural disasters that heavily impact the agriculture sector. A major flood in 2011 hit 18 of 24 provinces, affecting around 52,000 households (13 percent of the population). The next flood arrived in 2013 affecting six folds of the households size of the previous flood, followed by El-Nino events in 2015-2016. The Post Flood Early Recovery Assessment (PFERNA) recorded that the most common coping mechanisms are reducing expenses, relying on the assistance from NGO and government or taking loan from the MFIs \cite{kingdom}. 

One of the fundamental pillars of resilience framework is adaptive capacity \cite{iianalysing}. \cite{defiesta2014measuring} explicitly applied the adaptive capacity model in the Philippines by a composite index from the combination of survey data and experts' opinion that determines the weights of the indicators. \cite{phan2019gender} carried out the study of adaptive capacity with a gender lens in Vietnam. To the best of our knowledge, there is no quantitative study on resilience or adaptive capacity in Cambodia that has taken place to date. 

%Table 1
\begin{table}
  \caption{Adaptive capacity indicators}
  \label{table1}
  \centering
  \begin{tabular}{lll}
    \toprule
    %\multicolumn{2}{c}{Part}                   \\
    \cmidrule(r){1-2}
    Indicators   & Weights   & Proxy of sub-indicators   \\  %& Size ($\mu$m) \\
    \midrule
    Human resources & 0.072 & Age, schooled  \\%& %$\sim$100     \\
    Physical resources  & 0.231   & The loan is for agriculture  \\%& %$\sim$10      \\
    Financial resources  & 0.339   & \begin{tabular}{@{}l@{}}Number of loan accounts, proportion of local currency, \\ interest rate, loan term, loan balance\end{tabular} \\   
    Information   & 0.236  &    Urban * Female        \\
    Livelihood diversity  & 0.122  &    The number of income sources is more than two   \\
    \bottomrule
  \end{tabular}
\end{table}

%Figure 1
\begin{figure*}
\centering
\includegraphics[width=.95\linewidth]{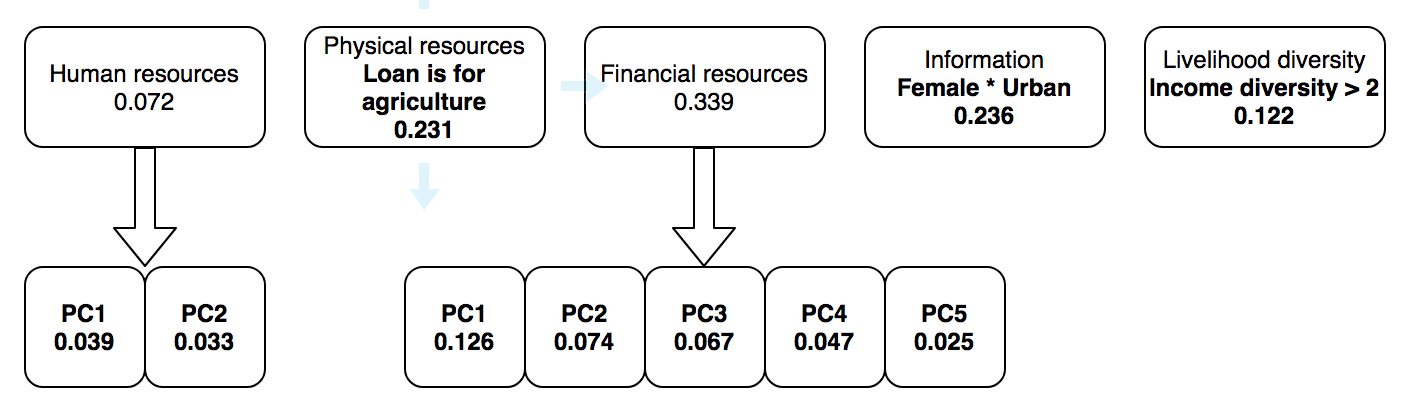}  
\caption{The weights of adaptive capacity variables, 2012}
\label{fig:diagram}

\end{figure*}

\section{Data}
\label{data}
The data for this study is generated from a business process of an anonymous MFI in Cambodia, accessed through a partnership with the UN Capital Development Fund (UNCDF). This is a longitudinal data with annual frequency between 2012-2015. The raw data is at account level  
that stores financial features such as annual loan balance, interest rate, loan term, and the choice of currency namely the Cambodia Riel (KHR) or foreign currencies, Thailand Bath (THB) and US Dollar (USD). The corresponding customer information is stored together at the account level to reflect the customer's background the first time they applied for a loan account. Some data might not reflect the true nature of time variant variables except for the first year of application. 

Extensive work on data pre-processing was largely conducted because there are several different versions of data entry style, even within a single year. Age was specifically corrected according to the year of birthday in order to maintain its time-variant nature. The last part of the pre-processing step is to aggregate the data to the customer level for each available year by taking the average. The final number of observations used for the analysis is close to 1.2 million.

\section{Method}
\label{method}
To enable us using the same indicators by \cite{defiesta2014measuring} and their associated weights, our first assumption is that this particular MFI in Cambodia targets rural societies that predominantly composed of agricultural communities, as which this is \cite{defiesta2014measuring}'s targeted observations. The second assumption is that those weights and likewise the agricultural livelihood are transferable across neighboring countries with similar characteristics. We further defined the proxy of sub-indicators according to the matched features, and if available, the previous literature like interaction between urban location and female networks to gather information \cite{phan2019gender}. The precise proxy sub-indicators used are reflected in Table \ref{table1}.

Proxy indicators with only one component automatically receive the experts' weight from \cite{defiesta2014measuring} such as physical resources, information and livelihood diversity. Proxy indicators with more than one components such as indicators of human resources and financial resources go through one further step to obtain the weight for each component as described in the next paragraph. 

Principal Component Analysis (PCA) was run to produce the standardized principal component scores for variables of both indicators separately, for each year. The principal components (PCs) become the new sets of proxy sub-indicators with their associated weights obtained from the proportion of variance explained. The final weights for the new proxy sub-indicators are generated from the multiplication between the experts' weights and the proportion of variance explained. Thus, the sum of the weights for each of the two indicators is being adjusted from one, as the total of proportion explained, to the proportions according to the experts' weights. Each variable now has the weight based on their status, either as the indicator or as the new proxy of sub-indicator. Figure \ref{fig:diagram} illustrates the variables and their associated weights in 2012, where bold fonts indicate their inclusion in calculating adaptive capacity index. The weights are always consistent for all indicators each year, but variant over time for the PCs. The individual adaptive capacity index is simply the summation of the weighted variables for each observation in a given year. The index was then aggregated at the district level for exploratory and cluster analyses.

Model based clustering analysis was performed at the individual and district level to capture patterns attributed to the pre, peri, and post 2013 flood with the mclust package in R \cite{scrucca2016mclust} that employs Gaussian finite mixture models. The best model is determined simultaneously with the best number of clusters using Bayesian Information Criteria (BIC). Further, key determinants of the clusters were selected through the combination of visual examination and decision tree. Insights from the comparison of substantial determinants between the lowest and the highest adaptability clusters might tell if any of those features can be improved by public policy interventions to develop resilience capacity.

\section{Discussion}
\label{discussion}

\subsection{Results}
\label{results}
\begin{figure*}
\centering
\begin{subfigure}{.45\textwidth}
  \includegraphics[width=.95\linewidth]{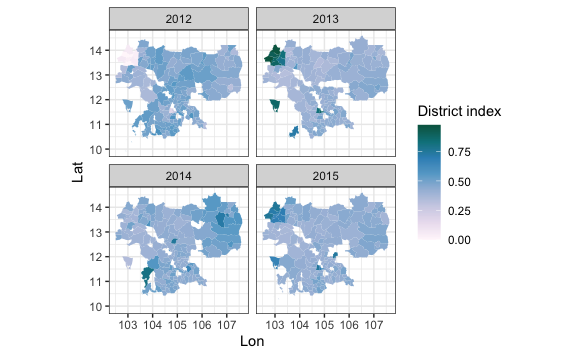}  
  \caption{Adaptive capacity in Cambodia}
  \label{fig:sub-a}
\end{subfigure}
\begin{subfigure}{.45\textwidth}
  \centering
  \includegraphics[width=.95\linewidth]{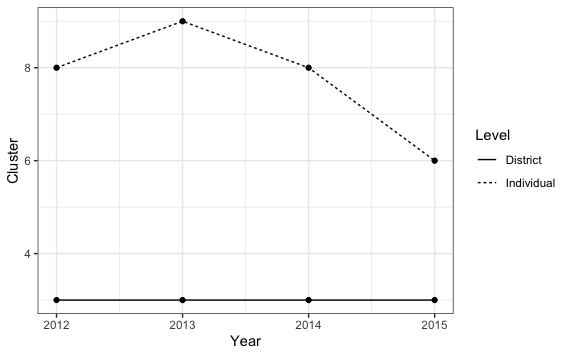}  
  \caption{Clustering adaptive capacity}
  \label{fig:sub-b}
\end{subfigure}

\caption{Exploratory analysis}
\label{fig:exploratory}

\end{figure*}

The exploratory analysis in Figure \ref{fig:exploratory} shows how the scoring results of adaptive capacity is further used to understand the dynamics of resilience in Cambodia. After aggregating at the district level then being pooled for all the years, Figure \ref{fig:sub-a} demonstrates the normalized index where a darker shade represents a higher score. There are districts that show contrast shades between non-disaster and disaster years, such as some areas in the northwest border and in the shore of Thailand Gulf (west border). They show a better capacity for adaptation during the disaster time. 

Although this snapshot is imperfect due to selection bias, as well as possibly not all adaptive capacity translates into adaptation, it offers some potential insights for disaster relief management and the longer term development projects. The clustering based classification in Figure \ref{fig:sub-b} additionally exhibits that the number of clusters at individual level reaches the peak in 2013. This is a potential insight of immediate divergence or inequality within the communities after the flood taking place. 

%Figure 2
%\begin{figure}[!tbp]
%  \centering
%  \begin{minipage}[b]{0.45\textwidth}
%    \includegraphics[width=\textwidth]{fig/map_district.png}
%    \label{fig:map}
%  \caption{Adaptive Capacity in Cambodia 2012-2015}
%  \end{minipage}
%  \hfill
%  \begin{minipage}[b]{0.45\textwidth}
%    \includegraphics[width=\textwidth]{fig/cluster_linechart.png}
%    \label{fig:cluster}
%  \caption{Clustering dynamics around the 2013 Cambodia flood}
%  \end{minipage}
%\end{figure}

%In Figure \ref{fig:test} we see a beautiful duck!
%\begin{figure}
%  \begin{minipage}{10cm}
%    blablabla
%  \end{minipage}
%  \begin{minipage}{10cm}
%    \includegraphics{ente}
%    \caption{test}
%\label{fig:test}
%  \end{minipage}
%\end{figure}
%\input{fig/cluster.tex}

\begin{figure*}
\centering
\begin{subfigure}{.45\textwidth}
  \includegraphics[width=.95\linewidth]{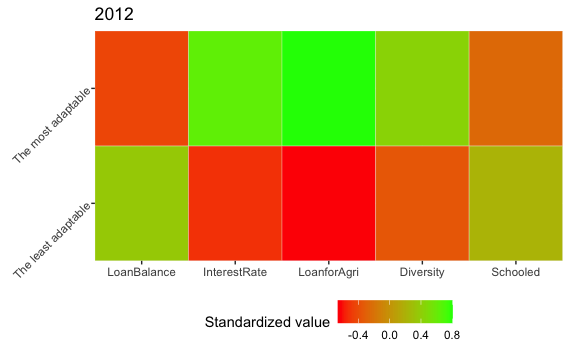}  
  \label{fig:heat-first}
\end{subfigure}
\begin{subfigure}{.45\textwidth}
  \centering
  \includegraphics[width=.95\linewidth]{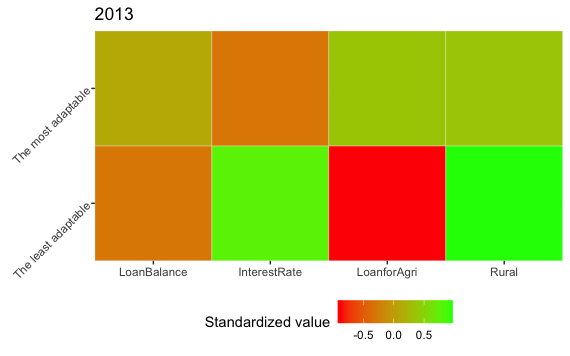}  
  \label{fig:heat-second}
\end{subfigure}

%\newline

\centering
\begin{subfigure}{.45\textwidth}
  \centering
  \includegraphics[width=.95\linewidth]{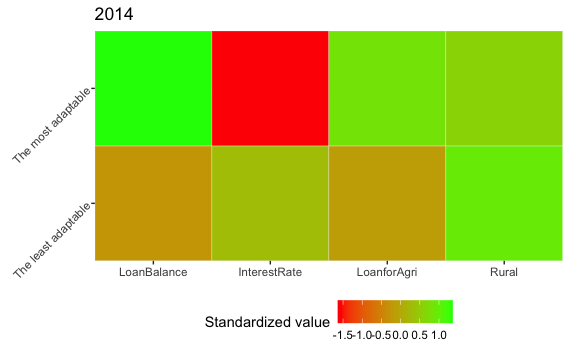}  
  %\caption{2014}
  \label{fig:heat-third}
\end{subfigure}
\begin{subfigure}{.45\textwidth}
  \centering
  \includegraphics[width=.95\linewidth]{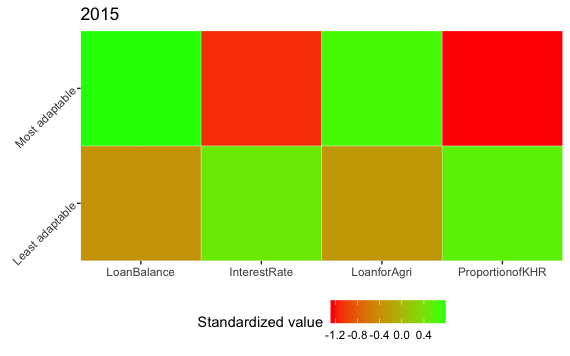}  
  %\caption{2015}
  \label{fig:heat-fourth}
\end{subfigure}

\caption{The heatmap of key determinants on adaptability}
\label{fig:heatmap}

\end{figure*}

Figure \ref{fig:heatmap} contrasts the vital characteristics between clusters with the highest and the lowest adaptive capacity over time to reveal the dynamics of adaptation ability determinants. Insights from these charts potentially offer an explanation on different coping strategies to respond to the disasters that could be useful to inform interventions. For instance, the most adaptable cluster shows a strong preference to borrow money for agriculture purposes and a general tendency to take loan with lower interest rate, while the least adaptable cluster is totally on the opposite side. Designing policy responses could capitalize this information by investigating further how those choices or situations are linked to improving resilience capacity. Another potential further examination is the individual considerations or the structural problems in those two aspects.

\subsection{Validation}
The validation of adaptive capacity index against two external data sources, Demographic and Health Survey (DHS) 2014 and Finscope 2015, was conducted via their matched variables. The results consistently show that customers in this particular MFI are narrowly segmented compared to the general population. The observations are skewed to those with formal education, female and concentrated at middle age. Results from Finscope offer an additional insight that the amount they borrowed is relatively smaller than that of the average population. 

\subsection{Limitation}
First, the unsupervised model used in this study limits the maximum number of clusters to be nine. The divergence in Figure \ref{fig:sub-b} might be more extreme without the limit. Second, the coverage is segmented therefore the results do not hold for the whole Cambodia population. Third, the data source comes from the financial sector known for their strict confidentiality, thus our approach might be hard for scaling up or reproducing in other settings. We hope that in the future, advanced techniques or strategies could create a nice balance between data privacy and data as public good/infrastructure, hence more data from the private sector could be used responsibly for a greater social impact.

\section{Concluding Remarks}
Vulnerability to natural disasters in many low income countries exacerbates their already devastating situation. Given the lack of sufficient public data infrastructure as well as typical budget constraint circumstances, suitable alternative data sources from the private sector is a potential quick solution to inform resilience planning and design via machine learning approach. In this paper, we show an exploratory study to understand the adaptive capacity in Cambodia with MFI data by using unsupervised learning methods. Such a study would help identify the baseline level of the ability for adaptation and advise the essential characteristics. These insights could inform a broader study about resilience or a more responsive public policy in humanitarian spaces.

%Figure 3

%\begin{figure*}
%\centering
%\begin{subfigure}{.45\textwidth}
%  \includegraphics[width=.95\linewidth]{fig/heatmap_ac12.png}  
%  %\caption{2012}
%  \label{fig:heat-first}
%\end{subfigure}
%\begin{subfigure}{.45\textwidth}
%  \centering
%  \includegraphics[width=.95\linewidth]{fig/heatmap_ac13.png}  
%  %\caption{2013}
%  \label{fig:heat-second}
%\end{subfigure}

%\newline

%\centering
%\begin{subfigure}{.45\textwidth}
%  \centering
%  \includegraphics[width=.95\linewidth]{fig/heatmap_ac14.png}  
%  %\caption{2014}
%  \label{fig:heat-third}
%\end{subfigure}
%\begin{subfigure}{.45\textwidth}
%  \centering
%  \includegraphics[width=.95\linewidth]{fig/heatmap_ac15.png}  
%  %\caption{2015}
%  \label{fig:heat-fourth}
%\end{subfigure}

%\caption{Heatmap of adaptive capacity key characteristics}
%\label{fig:heatmap}

%\end{figure*}

\subsection*{Acknowledgements}
Dr. Jong Gun Lee is acknowledged for pitching the initial research idea. Sriganesh Lokanathan is acknowledged for editorial support. George Hodge is acknowledged for peer support. UNCDF, in particular Robin Gravesteijn Ph.D, is acknowledged for the continuous supports on the data partnership. Demographic and Health Survey is acknowledged for the data access. Finally, the support of the Government of Australia in the funding aspect of this work is gratefully acknowledged. 

\bibliography{references}

\begin{thebibliography}{8}
\providecommand{\natexlab}[1]{#1}
\providecommand{\url}[1]{\texttt{#1}}
\expandafter\ifx\csname urlstyle\endcsname\relax
  \providecommand{\doi}[1]{doi: #1}\else
  \providecommand{\doi}{doi: \begingroup \urlstyle{rm}\Url}\fi

\bibitem[Defiesta et~al.(2014)Defiesta, Rapera, et~al.]{defiesta2014measuring}
G.~Defiesta, C.~Rapera, et~al.
\newblock Measuring adaptive capacity of farmers to climate change and
  variability: Application of a composite index to an agricultural community in
  the philippines.
\newblock \emph{Journal of Environmental Science and Management}, 17\penalty0
  (2), 2014.

\bibitem[FAO(2020)]{fao20}
FAO.
\newblock {Cambodia at a glance}, 2020.
\newblock URL
  \url{http://www.fao.org/cambodia/fao-in-cambodia/cambodia-at-a-glance/en/}.

\bibitem[Houston(2015)]{houston2015bouncing}
J.~B. Houston.
\newblock Bouncing forward: Assessing advances in community resilience
  assessment, intervention, and theory to guide future work, 2015.

\bibitem[II()]{iianalysing}
R.~II.
\newblock Analysing resilience for better targeting and action.

\bibitem[Manyena et~al.(2011)Manyena, O'Brien, O'Keefe, and
  Rose]{manyena2011disaster}
B.~Manyena, G.~O'Brien, P.~O'Keefe, and J.~Rose.
\newblock Disaster resilience: a bounce back or bounce forward ability?
\newblock \emph{Local Environment: The International Journal of Justice and
  Sustainability}, 16\penalty0 (5):\penalty0 417--424, 2011.

\bibitem[of~Cambodia(2014)]{kingdom}
K.~of~Cambodia.
\newblock {Post-Food Early Recovery Need Assessment Report}, 2014.

\bibitem[Phan et~al.(2019)Phan, Jou, and Lin]{phan2019gender}
L.~T. Phan, S.~C. Jou, and J.-H. Lin.
\newblock Gender inequality and adaptive capacity: The role of social capital
  on the impacts of climate change in vietnam.
\newblock \emph{Sustainability}, 11\penalty0 (5):\penalty0 1257, 2019.

\bibitem[Scrucca et~al.(2016)Scrucca, Fop, Murphy, and
  Raftery]{scrucca2016mclust}
L.~Scrucca, M.~Fop, T.~B. Murphy, and A.~E. Raftery.
\newblock mclust 5: clustering, classification and density estimation using
  gaussian finite mixture models.
\newblock \emph{The R journal}, 8\penalty0 (1):\penalty0 289, 2016.

\end{thebibliography}

\end{document}